\documentclass[pdflatex,sn-mathphys-num]{sn-jnl}% Math and Physical Sciences Numbered Reference Style 
\usepackage{graphicx}%
\usepackage{multirow}%
\usepackage{amsmath,amssymb,amsfonts}%
\usepackage{amsthm}%
\usepackage{mathrsfs}%
\usepackage[title]{appendix}%
\usepackage{xcolor}%
\usepackage{textcomp}%
\usepackage{manyfoot}%
\usepackage{booktabs}%
\usepackage{algorithm}%
\usepackage{algorithmicx}%
\usepackage{algpseudocode}%
\usepackage{listings}%
\usepackage{adjustbox} % 用于调整表格大小
\usepackage{caption}
\usepackage[section]{placeins}

\begin{document}

\title[Article Title]{Neural Networks Remember More: The Power of Parameter Isolation and Combination}


\author*[1]{\fnm{Biqing} \sur{Zeng}}\email{zengbiqing137@163.com}

\author[2]{\fnm{Zehan} \sur{Li}}\email{liiizer@163.com}

\author[1]{\fnm{Aladdin} \sur{Ayesh}}\email{aladdin.ayesh@abdn.ac.uk}

\affil[1]{\orgdiv{Aberdeen Institute of Data Science and Artificial Intelligence}, \orgname{South China Normal University}}

\affil[2]{\orgdiv{School of Artificial Intelligence}, \orgname{South China Normal University}}


\abstract{Catastrophic forgetting is a pervasive issue for pre-trained language models(PLMs) during continual learning, where models lose previously acquired knowledge when sequentially trained on a series of tasks. The model’s ability to remain old tasks is referred to as stability, while its adaptability to new tasks is called plasticity. Therefore, the key to solving this problem is to find a trade-off between the plasticity and stability of the model. 
To address this issue, in this paper, we propose a novel method to achieve a balance between model stability and plasticity, thereby mitigating catastrophic forgetting.
More specific, our proposed approach leverages parameter isolation and subsequent combination strategy. Initially, in training stage, the model adapts on each downstream task via parameter isolation method to prevent potential inference among different tasks. We then combine all trained parameters which containing acquired knowledge by the task arithmetic method and finally apply to the backbone model. Empirical evaluations on continual language learning benchmarks substantiate the effectiveness of our approach, revealing a marked enhancement over existing state-of-the-art approaches.}




\maketitle

\section{Introduction}\label{sec1}

Pre-trained Language Models(PLMs) have shown outstanding performance on a diverse range of downstream Natural Language Processing(NLP) tasks~\cite{ref1}. In real-world application, PLMs are often deployed in dynamic environment, necessitating continual adapting on the new data while preserving previous learned knowledge. However, an intractable issue known as \textit{catastrophic forgetting}~\cite{ref2} arises during continual learning, where models may drastically forget previously acquired knowledge when adapting to new tasks.

There is a series of work focus on mitigating catastrophic forgetting. For instance, rehearsal-based methods alleviate the problem through retraining the model with historical data which is cached during the previous learning process~\cite{ref3,ref4}. Nevertheless, as the model size grows, retraining the model multiple times is infeasible because of the expensive computational cost. Moreover, access to historical data may be restricted in some situations due to privacy and security. In addition to rehearsal-based methods, parameter isolation is another popular method for mitigating catastrophic forgetting, which alleviate potential inference among different tasks by allocating a separate set of parameters for each task~\cite{ref5,ref6}. However, conventional parameter isolation methods are only appropriate for task-incremental Continual learning problems because they require a task-id to select the proper modules during the testing phase.

Motivated by the above issues, our proposed method employs parameter isolation and combination strategy to replace conventional methods. More specifically, we employ Parameter-Efficient Fine-Tuning(PEFT) methods~\cite{ref7}(Adapter~\cite{ref8} and LoRA~\cite{ref9} in our experiments) to adapt downstream tasks, which introduce a small set of external parameters and only fine-tune these parameters while backbone model counterpart is kept frozen. Subsequently, after training on all downstream tasks, we use task arithmetic method~\cite{ref10} to integrate the knowledge that the model has obtained, thereby overcoming the limitation of previous parameter isolation methods that require task-id during testing stage. 
Besides mitigating catastrophic forgetting, we would like to facilitate knowledge transfer~\cite{ref6} among diverse tasks. In our work, we show that simply initializing the current PEFT modules with those of previous tasks effectively improves the knowledge transfer between learned and new tasks.

To verify the effectiveness of our proposed method, we conduct extensive experiments on standard continual learning benchmarks. The results show that our method not only outperforms existing rehearsal-based methods, but also improves upon previous state-of-the-art rehearsal-free methods. For example, the EPI method~\cite{ref11} perform an average accuracy of 76.3\% in the full-shot setting, while our approach obtains superior performance(77.2\%).

To summarize, our contributions are as follows:
\begin{itemize}
    \item We develop a novel method for mitigating catastrophic forgetting based on parameter isolation and combination strategy, and the effectiveness is validated through extensive experiments.
    \item Our proposed approach achieves satisfactory performance without the need to save historical data compared to the previous rehearsal-based methods, thereby reducing the storage and computational consumption.
    \item Compared with conventional parameter isolation methods, our method integrates all knowledge that model has acquired during previous learning process. Hence, our method is not necessitated by a task-id during testing, making it suitable for more than just task-incremental continual learning.
\end{itemize} 

\section{Related Works}\label{sec2}
\subsection{Continual Learning}
Continual learning is a scenario that the model learn from a stream of data over time, distinguishing it from traditional methods that train on the stationary dataset.
In continual learning, LLMs commonly encounter the problem of catastrophic forgetting. This occurs because as the model adapts to new tasks, its parameters tend to deviate from the optimal values that were previously established for old tasks. This phenomenon significantly affects the performance and reliability of models in practical applications, especially when dealing with multiple tasks and datasets~\cite{ref12}. To address this issue, various strategies have been proposed in the past. Here, we discuss three widely used methods, which include replay~\cite{ref13,ref14}, parameter regularization~\cite{ref15,ref16,ref17}, and parameter isolation~\cite{ref18,ref19,ref20} methods.

Replay, also known as rehearsal, is based on the idea of training models by supplementing the training data of current task with representative previous data~\cite{ref14}. However, these approaches are not without risk, as they may lead to privacy leakage. Further more, as the model scales up, there is a corresponding increase in the required storage and computing resources. 

Parameter regularization restricts the update of model weights through adding a regularization term to the loss function that penalizes large changes to the network's parameters~\cite{ref16}. Although these methods alleviate the problem of forgetting to some extent, they may also reduce the model's ability to adapt to new tasks~\cite{ref17}. 

Parameter isolation methods avoid interference between different tasks by assigning certain parts of the model exclusively to specific tasks~\cite{ref20}. However, these methods are only applicable to task-incremental learning scenarios as they often require a task-id to select the correct model when testing.

\subsection{Task Arithmetic}
In our study, we employ Task Arithmetic~\cite{ref10}, a groundbreaking approach to combines all parameters corresponding to each individual task after training. Task Arithmetic represents an innovative paradigm to guiding model behavior, focusing on the use of task vectors. The task vector specifies a direction within the weight space of a pre-trained model, and adjusting the model along this direction enhances its performance on the specific task. These vectors are obtained by subtracting the weights of the pre-trained model from those of a fine-tuned model. Subsequently, we can leverage simple arithmetic operations, termed task arithmetic, on task vectors to edit a model. For instance, by adding task vectors, we can combine diverse models to create a more effective multi-task model. 

\section{Methodology}
\begin{figure}
    \begin{center}
        \includegraphics[width=1\textwidth]{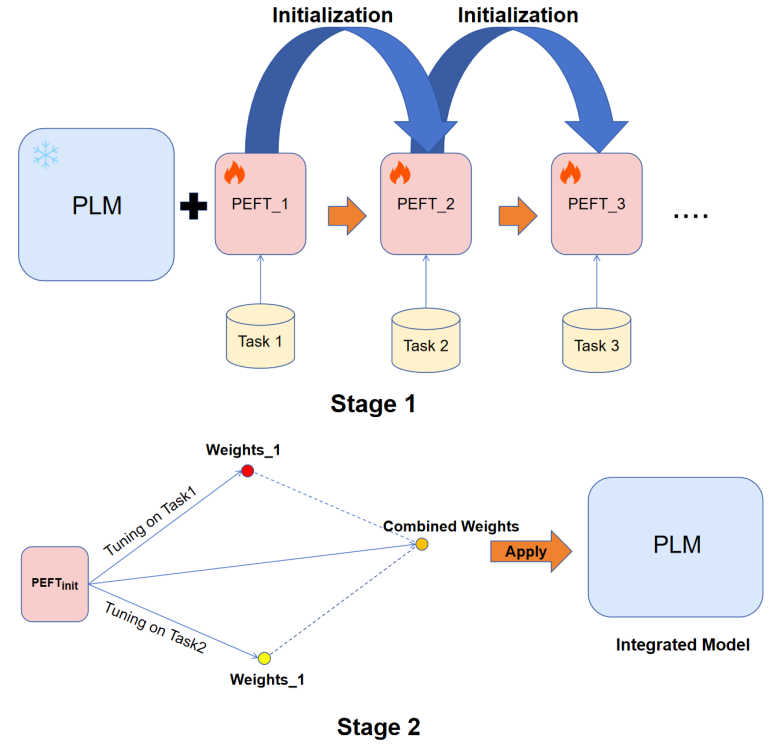}
        \caption{The process of our method. Our approach can be divided into two stages. In stage one, we train the model with PEFT method and initialize the next module with tuned weights. In stage two, as well as testing phase, we combine all adapted PEFT modules using Task Arithmetic method and subsequently apply to backbone model.} 
        \label{fig0}
    \end{center}
\end{figure}

Fig~\ref{fig0} summarises the approach presented in this paper. In the training stage, we assign a new PEFT module for each task and fine-tune model on training dataset while all of parameters of backbone model are kept frozen. After training on a task, we obtain and save the task vector by subtracting the initialization weights of the PEFT module from the tuned parameters. 

During testing time, we combine all acquired knowledge by adding task vectors according to task arithmetic method. Consequently, we apply the integrated task vector to the original pre-trained model and test its performance on each task's dataset. 

In the subsequent sections, we begin with an in-depth discussion of the parameter isolation with PEFT methods in Section 3.1. Moving forward, we illustrate the parameter combination method for testing in Section 3.2. Finally, Section 3.3 will delve into various knowledge transfer methods.  

\subsection{Parameter Isolation with PEFT}
One of the reasons for catastrophic forgetting in continual learning is the interference between tasks~\cite{ref21,ref22}. As the network parameters are adjusted to optimize the loss on the new task, they are often shifted away from their optimal values for the previously learned tasks. Therefore, a direct approach is to assign distinct parameters to each task, thereby preventing potential inference between them. However, assigning a distinct pre-trained model to each task would incur extremely high storage costs. Therefore we employ parameter-efficient fine-tuning~\cite{ref23} framework as an alternative. This approach adds a new component to the backbone model whenever a new task needs to be learned, while sharing the powerful pre-trained model across tasks. 

In the following, we briefly review Adapter and LoRA method. Adapter~\cite{ref8} are designed to make more general architectural adjustments, repurposing a pre-trained model for a specific downstream task. The adapter tuning strategy introduce a bottleneck structure to neutral network, including a couple of up/down project matrix. LoRA~\cite{ref9} allows us fine-tune a model indirectly by optimizing rank decomposition matrices of the dense layer's change during adaption. 
In a word, Adapter and LoRA utilize a shared pre-trained model, the weights denoted as $\Theta$, across various tasks. Additionally, they allocate a small number of external parameters for each task, represented as $\Phi_{n}$, where $n \in \{1, 2, \ldots, N\}$. The optimization objective for each task is of the form: 
\begin{equation}
    \Phi_n \leftarrow \underset{\Phi_n}{\text{argmin}} L_n(\Theta, \Phi_n)
\end{equation}
A pivotal feature of PEFT methods is their ability to effectively train models for specific tasks using a small set of external parameters integrated with a powerful PLM~\cite{ref23}. These approach not only significantly cut down the storage costs by parameter isolation methods, it also ensures high performance for individual tasks. 

\subsection{Parameter Combination with Task Arithmetic}
After training, we save the fine-tuned weights and subsequently calculate the task vectors. Denote a task vector $\tau$ for a specific task as:
\begin{equation}
    \tau_i = \Phi_i - \Phi_{\text{pre}}
\end{equation}
During testing, we combine all acquired knowledge by the following:
\begin{equation}
    \tau = \sum_{i=1}^{N} \lambda\tau_i\
\end{equation}
Where $\lambda$ is a scaling term and the optimal value is determined by validation dataset. Finally, we integrate all obtained task vectors into the original pre-trained model, resulting in a multi-task model proficient in all tasks, which is denoted as:
\begin{equation}
    \Theta \leftarrow \Theta_{\text{pre}} + \tau
\end{equation}

\subsection{Knowledge Transfer with Initialization}
Although completely separating parameters per task would eliminate any inference among tasks, it would also block the positive transfer among tasks~\cite{ref12}. Given a series of learned PEFT module $\Phi$ = \{$\Phi_1$, $\Phi_2$, \ldots, $\Phi_i$\}, we explore different strategies to leverage the knowledge acquired from previous tasks. Our goal is to enhance and accelerate the learning process for the current task. 

In our experiments, we demonstrate that initializing a new PEFT module with well-trained parameters is an effective strategy for facilitating knowledge transfer. A desirable starting point also significantly aids in accelerating the convergence of the training process for PEFT methods. As a result, we opt to initialize using previously learned modules instead of random initialization. In our experiments, we explore two initialization strategies for the PEFT modules. First, we initialize PEFT module with the parameters of the previous module, formalized as: 
\begin{equation}
    \Phi_i \leftarrow \Phi_{i-1}
\end{equation}
Second, we initialize the module using the average weights of the previously tuned modules, calculated as:
\begin{equation}
    \Phi_i \leftarrow \frac{1}{i-1} \sum_{t=1}^{i-1} \Phi_t
\end{equation}
The experimental results show that the effects of these two methods are similar(The former strategy is slightly better than the latter).

\section{Experiments}

\subsection{Setup}
\subsubsection{Datasets.}
Following the previous setting~\cite{ref4,ref25}, we assess the effectiveness of our approach using a well-established continual learning benchmark. The benchmark is composed of five diverse datasets: AG News (news), Yelp (business reviews), Amazon (product reviews), Yahoo!Answer (Q\&A), and DBPedia (encyclopedic articles). Collectively, these datasets are utilized to explore two primary text classification tasks: topic classification (AG News, Yahoo, and DBPedia) and sentiment classification (Yelp and Amazon). 

\subsubsection{Baselines.}
We employ the subsequent continual learning methods as baselines: 
\begin{itemize}
    \item \textbf{Fine-tune.} Fine-tuning the model with PEFT method on all tasks. Due to the severe degree of forgetting that occurs, this approach also represents the lower bound of continual learning.
    \item \textbf{Replay.} Following the conventional replay method, we cache some data during training process and re-train the model periodically.
    \item \textbf{MTL.} Training the model on all tasks simultaneously. Multi-task learning represents the upper bound of continual learning.
    \item \textbf{EPI~\cite{ref11}.} A parameter isolation method via prefix tuning~\cite{ref26} and using an non-parametric task identifier during testing. This method and ours leverage PEFT methods to implement the parameter isolation strategy similarly. However, our approach differs in that we create an integrated model by consolidating all weights of PEFT modules, which eliminates the need for task identification during the testing phase.
\end{itemize}

\subsubsection{Implement Details.} 
We utilize the Roberta-large~\cite{ref27} as the PLM in our experiments. We set the default adapter bottleneck to 32, default lora rank to 16 and the scaling term $\lambda$ to 0.25 while combing task vectors. 

\subsection{Main Results}
In Table.~\ref{tab1}, we compare the performance of our proposed method with established baselines across five benchmark datasets: AG News, Yelp, Amazon, Yahoo, and DBPedia. In our replay approach, we store 50 samples per class, which is equivalent to 2.5\% of the entire training dataset. The results presented in the table show that our approach achieves a new state-of-the-art (SOTA) result, outperforming previous methods. Furthermore, our approach, which does not utilize experience replay or task-ID, significantly reduces the gap to the upper bound(75.60\%). This demonstrates the effectiveness of our method in improving the performance of continual learning. Another key observation from the results is that the accuracy of our method remains stable across different orderings of the five datasets. This insensitivity to task sequence changes highlights the robustness and generalization capabilities of our approach. Overall, our method shows promise in improving the performance and stability of continual learning across various benchmark datasets.
\begin{table}[htbp]
    \renewcommand{\arraystretch}{1.2}
    \setlength{\tabcolsep}{8pt}
    \centering 
    \caption{Results on sampled setting of 5-datasets. The results are average over 2 runs. The best results of all methods are bolded. }
    \label{tab1}
    % 移除了scalebox，直接调整字体大小作为替代
    \normalsize % 或者使用\normalsize, \footnotesize等根据需要调整大小
    \begin{tabular}{l|ccc|c}
        \toprule
        Method & Order 1 & Order 2 & Order 3 & Avg \\
        \midrule
        FT & 27.4 & 26.7 & 33.1 & 29.0\\
        Replay & 52.1 & 67.3 & 61.7 & 60.3\\
        EPI & 73.1 & 72.5 & 72.4 & 72.7\\
        LoRA$_{16}$(\textbf{Ours}) & \textbf{73.8} & 73.4 & 74.4 & 73.8\\
        Adapter$_{32}$(\textbf{Ours}) & 73.7 & \textbf{74.2} & \textbf{74.7} & \textbf{74.2}\\
        \hline
        MTL & 75.6 & 75.6 & 75.6 & 75.6\\
        \bottomrule
    \end{tabular}
\end{table}

Table.~\ref{tab2} presents performance metrics across the full setting of five datasets. To evaluate and compare different methods, we have established two key methodological criteria, each denoted by a checkmark to signify its presence: Task-Identification (TI), which indicates the availability of task-id during inference; and Data Replaying (DR), which signifies whether the method necessitates the use of stored memory. The conclusions drawn from this analysis are largely in alignment with those from Table.~\ref{tab1}, but it also reveals additional insights. Specifically, our method, which forgoes rehearsal and does not utilize task-ID during the testing phase, surpasses previous state-of-the-art approach IDBR~\cite{ref25}, which relies on rehearsal. 
\begin{table}[htbp]
    \renewcommand{\arraystretch}{1.2}
    \setlength{\tabcolsep}{6pt}
    \centering 
    \caption{Results on full setting of 5-datasets. All results are based on the average of 2 independent runs.}
    \label{tab2}
    % 移除了scalebox，直接调整字体大小作为替代
    \normalsize % 或者使用\normalsize, \footnotesize等根据需要调整大小
    \begin{tabular}{l|cc|ccc|c}
        \toprule
        Method & TI & DR & Order 3 & Order 4 & Order 5 & Avg \\
        \midrule
        FT & & & 46.5 & & \\
        IDBR & & \checkmark & 75.4 & 75.7 & 75.8 & 75.6\\
        EPI & & & 76.5 & 76.4 & 76.0 & 76.3\\
        \hline
        LoRA$_{16}$(\textbf{Ours}) & & & 75.5 & 76.1 & 75.9 & 75.8\\
        Adapter$_{32}$(\textbf{Ours}) & & & \textbf{77.2} & \textbf{77.1} & \textbf{77.4} & \textbf{77.2}\\
        \bottomrule
    \end{tabular}
\end{table}

\subsection{Analysis}
To further inspect the proposed methods, we investigate the following research questions.
\subsubsection{Impact of module initialization on knowledge transfer} 
Ablation experiments were conducted to examine our proposed initialization method, and the results are presented in Table.~\ref{tab3}. The experiments are implemented with 2 different datasets order. The experimental results show that, compared to not implement initialization, our method significantly improves the model’s performance. Besides, we could find that the effect of these two methods are similar. Specifically, the "pre initialization" strategy is slightly better than the "mean initialization".
\begin{table}
    \renewcommand{\arraystretch}{1.2}
    \setlength{\tabcolsep}{60pt}
    \centering
    \caption{Results of ablation experiments focusing on the impact of initialization methods on model performance. \textbf{noinit} refers to the baseline results obtained without specialized initialization. \textbf{pre} indicates that the PEFT modules were initialized with the last PEFT module during the training process. \textbf{mean} signifies that the PEFT modules were initialized using the mean weights of the tuned modules.}
    \label{tab3}
    \normalsize
    \begin{tabular}{c|c}
        \toprule
        Method & Avg \\
        \midrule
        LoRA$_{noinit}$ & 22.0\\
        LoRA$_{pre}$ & \textbf{75.5}\\
        LoRA$_{mean}$ & 74.6\\
        \hline
        Adapter$_{noinit}$ & 30.0\\
        Adapter$_{pre}$ & \textbf{77.2}\\
        Adapter$_{mean}$ & 76.2\\
        \bottomrule
    \end{tabular}
\end{table}

\begin{figure}[htbp]
    \begin{center}
        \includegraphics[width=0.8\textwidth]{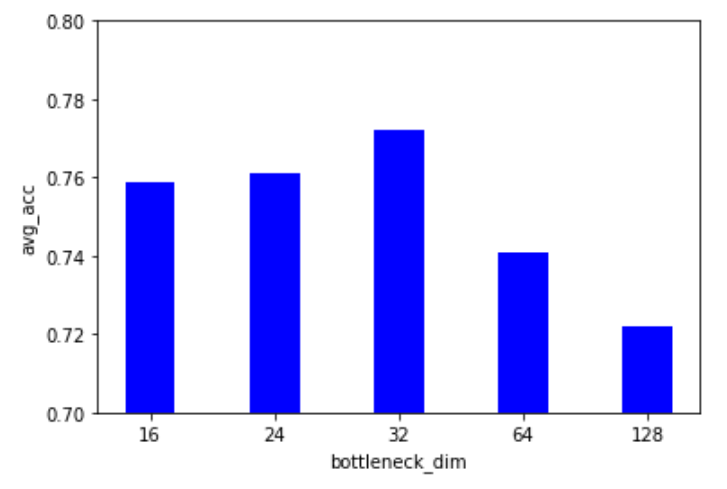}
        \caption{Average results of the Adapter method with varying bottleneck dimensions in the full setting, across five different datasets.(ag news $\rightarrow$ yelp $\rightarrow$ amazon $\rightarrow$ yahoo $\rightarrow$ db)} 
        \label{fig1}
    \end{center}
\end{figure}

\begin{figure}[htbp]
    \begin{center}
        \includegraphics[width=0.8\textwidth]{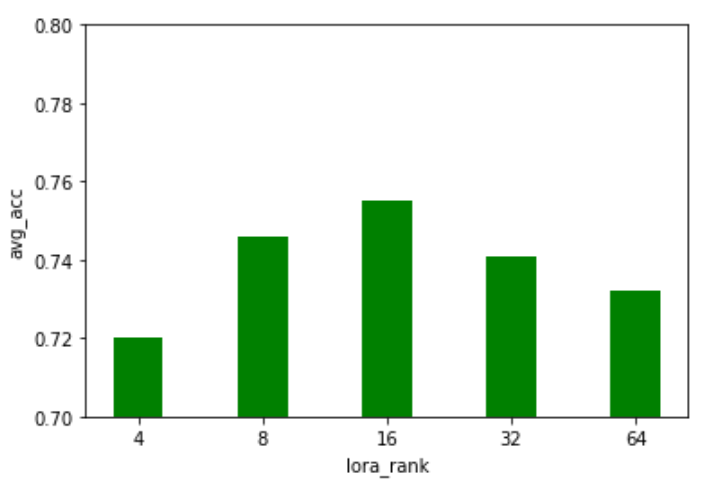}
        \caption{Mean results of the LoRA method with different LoRA ranks in the full setting, across five different datasets.(ag news $\rightarrow$ yelp $\rightarrow$ amazon $\rightarrow$ yahoo $\rightarrow$ db)}
        \label{fig2}
    \end{center}
\end{figure}

\subsubsection{Influence of different PEFT setting on model performance} 
To investigate the performance of different PEFT setting, we implement the experiments with various bottleneck dimension(for Adapter method) and LoRA rank(for LoRA method) to compare their effect. Fig.~\ref{fig1} and Fig.~\ref{fig2} presents average accuracy across the full setting of five datasets with different PEFT method settings. The similar results show that moderate dimension size more closer to the stability-plasticity trade-off of the model. We speculate that an excessively large dimension may diminish the plasticity of the model, whereas a dimension that is too small may result in lower stability of the model. 

\section{Conclusion}
In summary, in our work, we propose a novel method that leverages parameter isolation and combination to mitigate catastrophic forgetting for pre-trained language model(PLM), with its efficacy supported by extensive experimental evidence. Our approach exhibits superior performance to previous methods without the requirement for storing historical data and task identification, leading to reduced storage and computational costs. Besides, we introduce two initialization method to facilitate the knowledge transfer between tasks.

\section*{Declarations}
\begin{itemize}
    \item Competing Interests. No potential conflict of interest was reported by the authors.
    \item Data Availability Statements. The data that support the finding of this study are openly available in Hugging Face repository at https://huggingface.co/.
    \item Funding. This study was funded by National Natural Science Foundation of China Research Project(grant number 62076103); Guangdong basic and applied basic research project(grant number 2021A1515011171); Guangzhou basic research plan, basic and applied basic research project(grant number 202102080282).
\end{itemize}

\bibliography{references}% common bib file

\end{document}